\documentclass{bmvc2k}
\usepackage{comment}
\usepackage{tabularx}
\usepackage{ragged2e}
\usepackage{booktabs}       
\usepackage{amssymb}
\usepackage{wrapfig}
\usepackage{float} 

\newcolumntype{Y}{>{\RaggedRight\arraybackslash}X}
\usepackage{adjustbox}

\usepackage{multirow}

\newcommand{\etal}{\textit{et al}. }
\newcommand{\ie}{\textit{i}.\textit{e}., }
\newcommand{\eg}{\textit{e}.\textit{g}. }
\input{macros}

\title{Zero-Shot Sign Language Recognition: Can Textual Data Uncover Sign Languages?}

\addauthor{Yunus Can Bilge}{yunuscan.bilge@hacettepe.edu.tr}{1}
\addauthor{Nazli Ikizler-Cinbis}{nazli@cs.hacettepe.edu.tr}{1}
\addauthor{Ramazan Gokberk Cinbis}{gcinbis@ceng.metu.edu.tr}{2}

\addinstitution{
 Hacettepe University\\
 Department of Computer Engineering \\
 Ankara, Turkey
}
\addinstitution{
 Middle East Technical University\\
 Department of Computer Engineering\\
 Ankara, Turkey
}

\runninghead{Bilge et al.}{Zero-Shot Sign Language Recognition}

\def\eg{\emph{e.g}\bmvaOneDot}

\def\etal{\emph{et al}\bmvaOneDot}

\begin{document}

\maketitle

\begin{abstract}
We introduce the problem of zero-shot sign language recognition (ZSSLR), where the goal is to leverage
models learned over the seen sign class examples to recognize the instances of unseen signs. To this end, we propose to utilize the
readily available descriptions in sign language dictionaries as an intermediate-level semantic representation
for knowledge transfer. We introduce a new benchmark dataset called \textit{ASL-Text} that consists of 250
sign language classes and their accompanying textual descriptions. Compared to the ZSL datasets in other domains
(such as object recognition), our dataset consists of limited number of training examples for a large number of
classes, which imposes a significant challenge.  We propose a framework that operates over the body and hand regions
by means of 3D-CNNs, and models longer temporal relationships via bidirectional LSTMs. By leveraging the descriptive
text embeddings along with these spatio-temporal representations within a zero-shot learning framework, we show that
textual data can indeed be useful in uncovering sign languages. We anticipate that the introduced approach and the
accompanying dataset will provide a basis for further exploration of this new zero-shot learning problem.
\end{abstract}
\section{Introduction}
\label{sec:intro}

Sign language recognition (SLR) is one of the open problems in computer vision, with several challanges remain to be addressed. For instance, while the definitions of the signs are typically clear and structural, the meaning of a sign can change based on the shape, orientation, movement, location of the hand, body posture, and non-manual features like facial expressions \cite{wu1999vision,stokoe2005sign}. 
Even the well-known hand shapes in the sign languages can be difficult to discriminate and annotate even due to viewpoint changes~\cite{neidle2012challenges}. In addition, similar to natural languages, sign languages change and embrace variations over time. 
Therefore, the development of scalable methods to deal with such variations and challenges is needed. 

The existing SLR approaches typically require a large number of annotated examples for each  sign class of interest~\cite{cihan2018neural,camgoz2017subunets,koller2015continuous,koller2016deep,stoll2018sign}.
Towards overcoming the annotation bottleneck in scaling up SLR recognition, we explore the idea of recognizing sign language classes with no annotated visual examples, by leveraging their textual descriptions. To this end, we introduce the problem of \textit{zero-shot sign language recognition} (ZSSLR). Unlike the traditional supervised SLR, where training and test classes are the same, in ZSSLR, the aim is to recognize sign classes that are not seen during training. 
Compared to the commonly studied ZSL problems, where most seen (training) classes have a large number of per class samples \cite{patterson2012sun, wah2011caltech, lampert2014attribute, farhadi2009describing}, ZSSLR takes ZSL into a new extreme where most seen classes have only few training examples. This challenging situation corresponds to a \textit{hard zero-shot learning} problem~\cite{madapana2018hard}. 

In order to realize seen to unseen class transfer, we use textual descriptions of sign classes taken from a sign
language dictionary. Using a sign language dictionary for obtaining the class representations has two major advantages of being
(i) readily available, and, (ii) prepared by the sign language experts in a detailed way.
In this manner, we construct our ZSSLR approach on highly-informative class descriptions without requiring any ad-hoc  
annotations, unlike possible alternative approaches such as the attribute-based ZSL~\cite{lampert2014attribute,farhadi2009describing}.

\begin{figure}[t]
\centering
    \includegraphics[width=\textwidth]{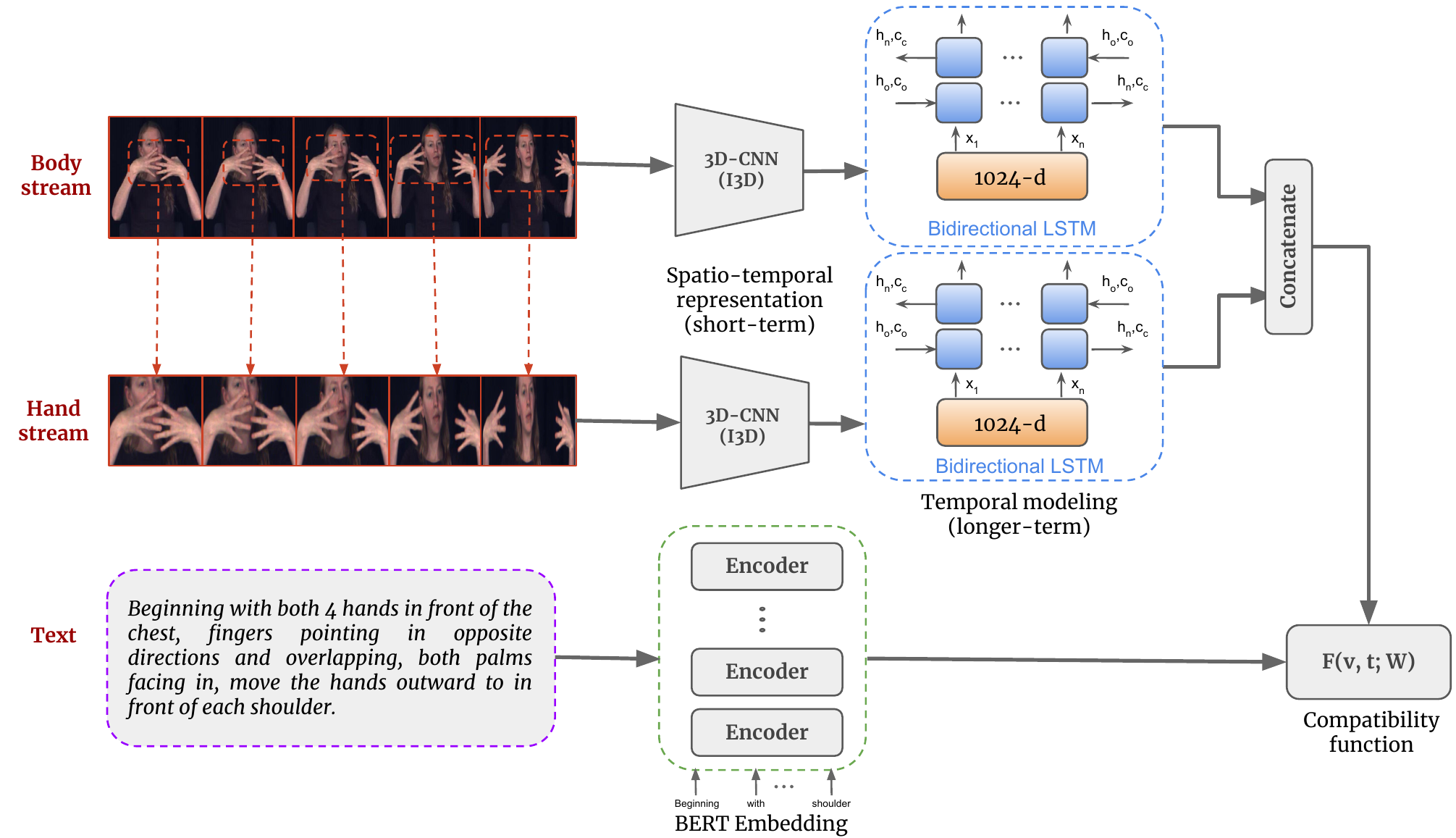}
   
   \vspace{-1mm}
   \caption{Overview of the proposed zero-shot sign language recognition (ZSSLR) approach. }

    \label{fig:main_figure}
\end{figure}

To study ZSSLR, we introduce a new benchmark dataset with 250 sign classes and their textual definitions. Our benchmark dataset is constructed from the ASLLVD corpus \cite{neidle2012challenges}, where the top 250 sign classes with most in-class variance are selected and the corresponding descriptions are gathered from the Webster American Sign Language Dictionary \cite{costello1999random}\footnote{The dataset is available at: \url{https://ycbilge.github.io/zsslr.html}}.

We propose an embedding-based framework for ZSSLR that consists of two main components. First component models the visual data with extra attention to temporal and spatial structure, using 3D-CNNs and LSTMs. These networks operate over body and hand regions of video in conjunction, since hands are important focal points of signs. The second component, the ZSL component, learns an embedding of the visual representation to the closest text description. We rigorously evaluate our proposed approach on ASL-Text dataset and show its advantages.

To summarize, the main contributions of the paper are as follows: (i) we formulate the problem of zero-shot sign language recognition (ZSSLR),
(ii) we define a new benchmark dataset for ZSSLR called ASL-Text,
(iii) we propose a spatio-temporal representation that focuses hand and full body regions via 3D-CNNs and LSTMs and learn it in an end-to-end manner for ZSSLR, and, (iv) we present the benchmark results with detailed analyses.
\section{Related Work}

\noindent\textbf{Sign Language Recognition (SLR).} SLR has been studied more than three decades~\cite{tamura1988recognition}. The mainstream SLR approaches can be grouped into two categories: (i) Isolated SLR~\cite{wang2016isolated}, and, (ii) Continuous SLR~\cite{cihan2018neural}. Our work belongs to isolated SLR category as we target to recognize individual sign instances. 

Early SLR methods mostly use hand-crafted features in combination with a classifier, such as support vector machines. Hidden Markov Models (HMM), Conditional Random Fields and neural network based approaches have also been explored to model the temporal patterns~\cite{grobel1997isolated, huang1998sign}. Recently, several deep learning based SLR approaches have been proposed~\cite{huang2015sign, pigou2016sign, molchanov2016online, koller2016deep, camgoz2017subunets, cui2017recurrent, stoll2018sign, cihan2018neural, narayana2018gesture}.  

Despite the relative popularity of the topic, the problem of annotated data sparsity has been seldomly addressed in SLR research. Farhadi and Forsyth \cite{farhadi2006aligning} is first to study the alignment of sign language video subtitles and signs in action to overcome annotation difficulty. Their approach~\cite{farhadi2007transfer} was based on transferring large amounts of labelled avatar data with few labelled human signer videos to spot words in videos. Buehler \etal \cite{buehler2009learning} also try to reduce the annotation effort by using the subtitles of British Sign Language TV broadcasts. They apply Multiple Instance Learning (MIL) to recognize signs out of TV broadcast subtitles. Kelly \etal \cite{kelly2011weakly} and Pfister \etal \cite{pfister2013large} also use subtitles of TV broadcasts. Pfister \etal \cite{pfister2013large} differ from the two aforementioned MIL studies 
as they track the co-occurrences of lip and hand movements to reduce the search space for visual and textual content mapping. Nayak~\etal \cite{nayak2009automated} proposes to locate signs in continuous sign language sentences using iterated conditional modes. 
Pfister \etal \cite{pfister2014domain} define each sign class with one strongly supervised example, and, train an SVM based detector out of one-shot examples. The resulting detector is then used to acquire more training samples from an another weakly-labeled data. Koller \etal \cite{koller2016deep} propose a combined CNN and HMM approach to train a model with large but noisy data. None of the aforementioned models approach the problem of annotated data sparsity from a zero-shot learning perspective. 

\noindent\textbf{Zero-Shot Learning.}
ZSL has been a focus of interest in vision and learning research in recent years, especially following the pioneering works of Lampert \etal \cite{lampert2009learning} and Farhadi \etal \cite{farhadi2009describing}. The main idea is learning to generalize a recognition model for identifying unseen classes. Most of the ZSL approaches rely on transferring semantic information from seen to unseen classes. For this purpose, semantic attributes are largely used in the literature \cite{ferrari2008learning, farhadi2009describing, liu2011recognizing, parikh2011relative, fu2014learning, lampert2014attribute, jain2015objects2action}. Semantic word/sentence vectors and concept ontologies are also studied in this context \cite{rohrbach2011evaluating, norouzi2013zero, elhoseiny2013write, frome2013devise, socher2013zero, mensink2014costa, lei2015predicting}. Label embedding models are explored to make connection between seen and unseen classes via semantic representations \cite{akata2013label,frome2013devise, norouzi2013zero, fu2014transductive, romera2015embarrassingly, qin2017zero, sumbul2018fine}. Akata \etal \cite{akata2013label} propose a method to learn a compatibility function from visual to semantic feature space. 
As opposed to models that learn to map to a semantic space, there are also studies that learn to map to a common embedding space \cite{romera2015embarrassingly, fu2014transductive}.

Recently, ZSL has also been explored in the context of action recognition. Liu \etal \cite{liu2011recognizing} is first to propose attribute based model for recognizing novel actions. Jain \etal \cite{jain2015objects2action} propose a semantic embedding based approach using commonly available textual descriptions, images, and object names. Xu \etal \cite{Xu2015SemanticES} propose a regression based method to embed actions and labels to a common embedding space. 
Xu \etal \cite{xu2017transductive} also use word-vectors as a semantic embedding space in transductive settings. Wang \etal \cite{wang2017alternative} exploit human actions via related textual descriptions and still images. Their aim is to improve word vector semantic representations of human actions with additive information. Habibian \etal \cite{habibian2017video2vec} also propose to learn semantic representations of videos with freely available video and relevant descriptions.  
Qin \etal \cite{qin2017zero} use error-correcting output codes to overcome the disadvantages of attributes and/or semantic word embeddings for information transfer. Compared to action recognition, in SLR, even a subtle change in motion and/or handshape can change the entire meaning. Therefore, we argue that specialized methods are required for zero-shot recognition in SLR. 

There are a couple of recent methods that introduce ZSL to gesture recognition. However, these methods are mostly limited to either robot interactions with single held-out classes (\cite{thomason2016recognizing}), or based on attributes with limited datasets (\cite{madapana2018hard}). We argue that attribute based semantic representations can be subjective and there is a high chance of missing beneficial attributes when annotating attributes manually. As also noted by \cite{zhu2018towards}, attribute based semantic representations are difficult to scale up as defining the attributes of even a single class can require a laborious amount of human effort. In this work, we work over an extensive dataset of classes for sign language and present an approach that does not require any manual attribute annotations. 
\begin{figure*}[!t]
\centering
\begin{tabular}{>{\raggedright\arraybackslash}p{0.235\linewidth}@{$\;\;$}>{\raggedright\arraybackslash}p{0.235\linewidth}@{$\;\;$}>{\raggedright\arraybackslash}p{0.235\linewidth}@{$\;\;$}>{\raggedright\arraybackslash}p{0.235\linewidth}}
\toprule
\scriptsize{\textbf{BICYCLE}} & 
\scriptsize{\textbf{LIKE}} & 
\scriptsize{\textbf{ALONE}} & 
\scriptsize{\textbf{FAMILY}} \\
\includegraphics[width=0.25\linewidth, height=1.5cm]{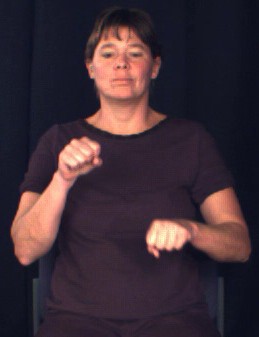}
\includegraphics[width=0.25\linewidth, height=1.5cm]{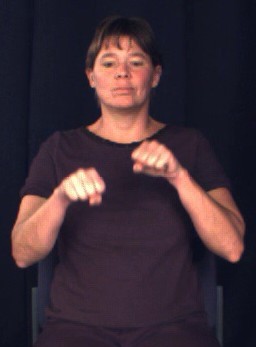}
\includegraphics[width=0.25\linewidth, height=1.5cm]{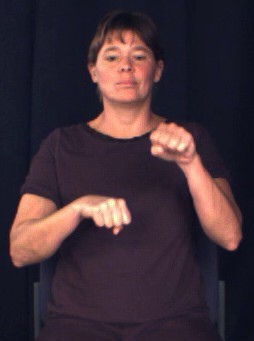} 
\begin{minipage}{0.23\textwidth}
\tiny{Move both \textcolor{red}{S} hands in alternating forward circles, palms facing down, in front of each side of the body.\\}
\end{minipage} 
 & 
\includegraphics[width=0.25\linewidth, height=1.5cm]{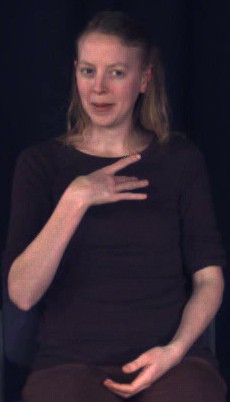}
\includegraphics[width=0.25\linewidth, height=1.5cm]{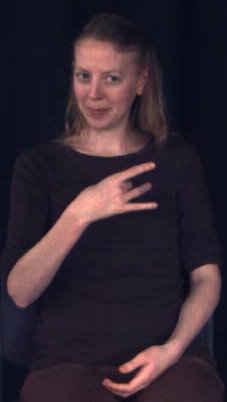}
\includegraphics[width=0.25\linewidth, height=1.5cm]{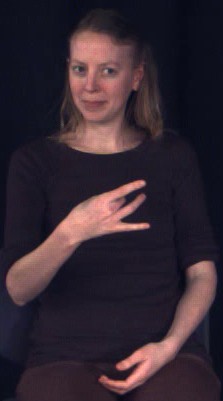}

\begin{minipage}{0.23\textwidth}
\tiny{Beginning with the bent thumb and middle finger of the right \textcolor{red}{5} hand touching the chest, palm facing in, bring the hand forward while closing the fingers to form an \textcolor{red}{8} hand.\\}
\end{minipage} 
 & 
 \includegraphics[width=0.25\linewidth, height=1.5cm]{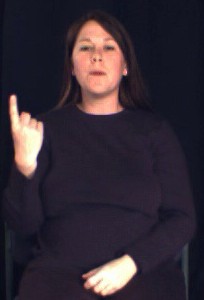}
\includegraphics[width=0.25\linewidth, height=1.5cm]{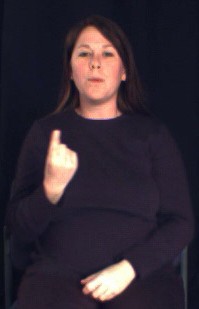}
\includegraphics[width=0.25\linewidth, height=1.5cm]{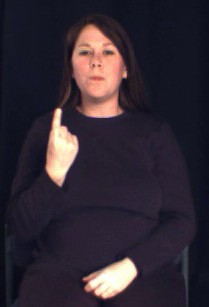}
\begin{minipage}{0.23\textwidth}
\tiny{With the right index finger extended up, move the right hand, palm facing back, in a small repeated circle in front of the right shoulder.\\}
\end{minipage} 
& 
\includegraphics[width=0.25\linewidth, height=1.5cm]{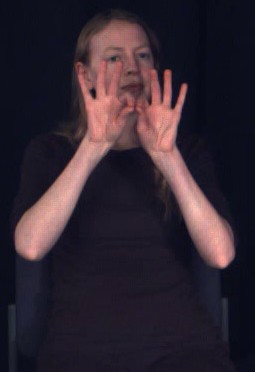}
\includegraphics[width=0.25\linewidth, height=1.5cm]{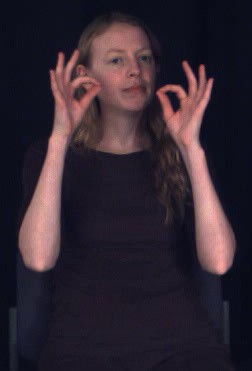}
\includegraphics[width=0.25\linewidth, height=1.5cm]{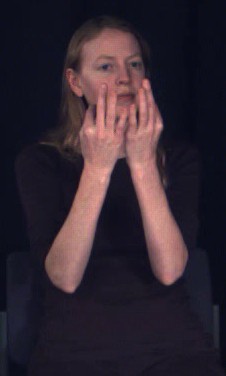}
\begin{minipage}{0.23\textwidth}
\tiny{Beginning with the fingertips of both  \textcolor{red}{F} hands touching in front of the chest, palms facing each other, bring the hands away from each other in outward arcs while turning the palms in, ending with the little fingers touching.}
\end{minipage}
\\
\scriptsize{\textbf{EAT}} & 
\scriptsize{\textbf{HIGH}} & 
\scriptsize{\textbf{HIT}} & 
\scriptsize{\textbf{LIBRARY}} \\

\includegraphics[width=0.25\linewidth, height=1.5cm]{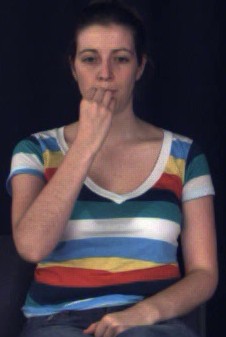}
\includegraphics[width=0.25\linewidth, height=1.5cm]{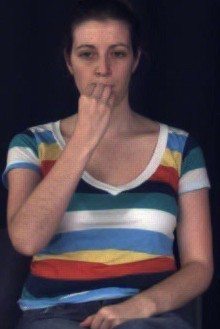}
\includegraphics[width=0.25\linewidth, height=1.5cm]{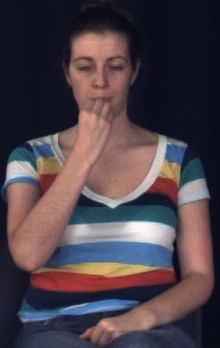}
\begin{minipage}{0.23\textwidth}
\tiny{Bring the fingertips of the right flattened \textcolor{red}{O} hand, palm facing in, to the lips with a repeated movement.}
\end{minipage}
 & 
\includegraphics[width=0.25\linewidth, height=1.5cm]{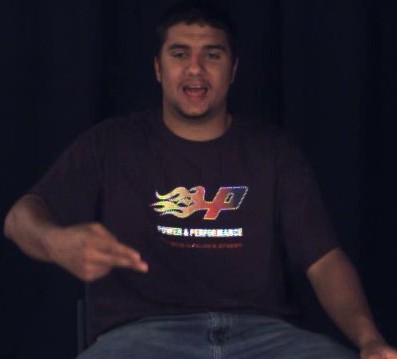}
\includegraphics[width=0.25\linewidth, height=1.5cm]{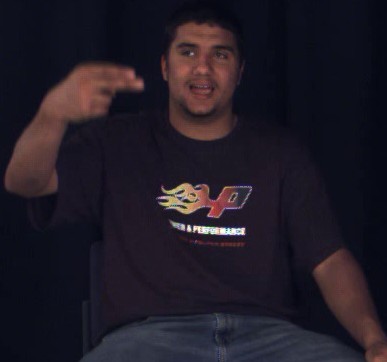}
\includegraphics[width=0.25\linewidth, height=1.5cm]{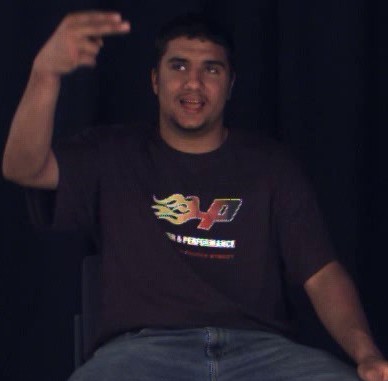}
\begin{minipage}{0.23\textwidth}
\tiny{Move the right \textcolor{red}{H} hand, palm facing left and fingers pointing forward, from in front of the right side of the chest upward to near the right side of the head.}
\end{minipage} 
 & 
 \includegraphics[width=0.25\linewidth, height=1.5cm]{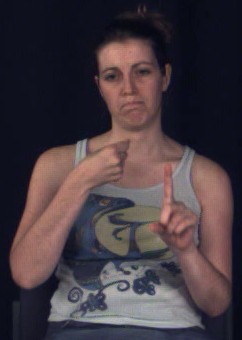}
\includegraphics[width=0.25\linewidth, height=1.5cm]{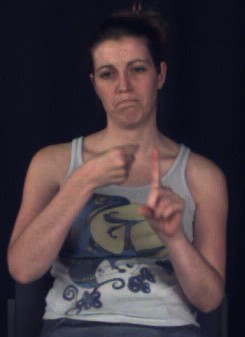}
\includegraphics[width=0.25\linewidth, height=1.5cm]{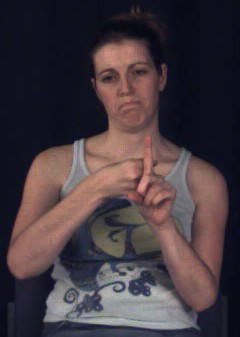}
\begin{minipage}{0.23\textwidth}
\tiny{Strike the knuckles of the right \textcolor{red}{A} hand, palm facing in, against the extended left index finger held up in front of the chest, palm facing right.}
\end{minipage} 
& 
\includegraphics[width=0.25\linewidth, height=1.5cm]{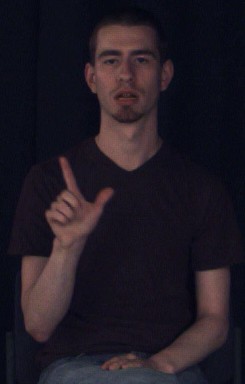}
\includegraphics[width=0.25\linewidth, height=1.5cm]{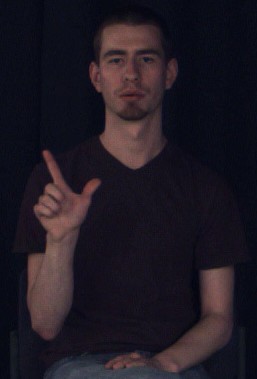}
\includegraphics[width=0.25\linewidth, height=1.5cm]{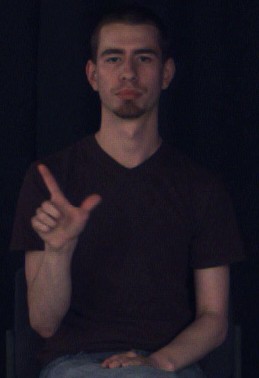}

\begin{minipage}{0.23\textwidth}
\tiny{Move the right \textcolor{red}{L} hand, palm facing forward, in a circle in front of the right shoulder.}
\end{minipage}\\

\bottomrule
\end{tabular}
\caption{Example sequences and corresponding textual descriptions from the  ASL-Text dataset. For visualization purposes, only the person regions of the videos are shown. }
\label{fig:dataset}
\vspace{-5mm}
\end{figure*}

\section{ASL-Text Dataset}
To facilitate ZSSLR research, we use the ASLLVD dataset \cite{neidle2012challenges}, which
is the largest isolated sign language recognition dataset available, to the best of our knowledge.
We select top 250 sign classes, ranked by the number of samples per class, from ASLLVD signer variances and augment this dataset with the textual definitions of the signs from Webster American Sign Language Dictionary \cite{costello1999random}. We refer to this new benchmark dataset as \textit{ASL-Text}. Example frames and their textual descriptions for the ASL-Text dataset are presented in Figure~\ref{fig:dataset}.

The textual descriptions include the detailed instructions of a sign with emphasis on four basic parts: hand-shape, orientation of the palms (forward, backward, etc.), movements of the hands (right, left, etc.), and the location of hands with respect to the body (in front of the chest, each side of the body, right shoulder, etc.). Moreover, some descriptions also include non-manual cues such as the facial expressions, head movements and body posture. Hand shapes are described with specialized vocabulary including the terms {\em F-hand}, {\em A-hand}, {\em S-hand}, {\em 5-hand}, {\em 8-hand}, {\em 10-hand}, {\em open-hand}, {\em bent-v hand}, {\em flattened-o hand} \cite{costello1999random}. Such a specialized vocabulary highlights the fact that ASL is a {\em language} on its own.  From the example hand shapes shown in Figure~\ref{fig:dataset}, it can be seen that the textual sign language descriptions are indeed quite indicative of the ongoing gesture. 

In the ASL-Text dataset, there are 1598 videos (54151 frames) in total for the 250 sign classes. 
The number of frames of individual videos range between 6 to 116, where the average sequence length is 33 frames. For ZSL purposes, we split the dataset into three disjoint sets (train, validation and test) based on classes. Train set includes 170, validation and test sets include 30 and 50 disjoint classes, respectively. The classes with most signer variation and in-class samples are assigned to training set. The remaining classes, which have relatively lower number of visual examples, are allocated into validation and test sets. This is done to demonstrate the real-world case; \ie it is harder to train classifiers for classes that are rarely seen, therefore, we train with the classes that have relatively more examples and test on the rare classes.
Overall, we have 1188, 151, and 259 video samples in training, validation, and test sets. The average length of the textual descriptions is 30 words per description, where the total vocabulary includes 154 distinct words.

The average number of instances per class is 7 for the training classes and 5 for the validation and test classes. Note that, still, the number of examples per class even for training is considerably lower than the commonly studied ZSL datasets, \eg AWA-2 \cite{xian2017zero} and SUN Attribute \cite{patterson2012sun},
on which hundreds of per-class examples are used for training.

\section{Methodology}
In this section, we first give a formal definition of the problem, and then explain the components of the proposed
approach, an overview of which is given in Figure~\ref{fig:main_figure}. The implementation details can be found in
Section~\ref{sec:implementationdetails}. 

\newcommand{\domainv}{\mathbb{V}} 
\newcommand{\domaint}{\mathbb{T}} 
\newcommand{\clss}{\mathbb{C}_\text{s}} 
\newcommand{\clsu}{\mathbb{C}_\text{u}} 
\newcommand{\loss}{\ell} 

\myparagraphwithspace{Problem definition.}
In ZSSLR, there are two sources of information: (i) the
\textit{visual domain} $\domainv$, which consists of sign videos, and, (ii) the \textit{textual domain} $\domaint$, which
includes the textual sign descriptions.  At training time, the videos, labels and the sign descriptions,
are available only for the \textit{seen} classes, $\clss$. At test time, our goal is to correctly
classify the examples of novel {\em unseen} classes, $\clsu$, which are distinct from the seen classes.

The training set $S_{tr} = \{ (v_{i}, c_i )\}_{i=1}^{N}$ consists of $N$ samples where $v_{i}$ is the $i$-th training video and
$c_i \in \clss$ is the corresponding sign class label.
We assume that we have access to a textual description of each class $c$, represented by $\tau(c)$.
The goal is to learn a zero-shot classifier that can correctly assign 
each test video to a class in $\clsu$, based on the textual descriptions.

In our approach, we aim to construct a label embedding based zero-shot classification model. For this purpose, we define
the compatibility function $F(v,c)$ as a mapping from an input video and class pair to a score representing the
confidence that the input video $v$ belongs to the the class $c$. Given the compatibility function $F$, the test-time
zero-shot classification function $f: \domainv \rightarrow \clsu$ is defined as:
\begin{equation}
f(v) = \argmax_{c \in \clsu} F(v,c) .
\label{eq:clsfunc}\end{equation}
In this way, we leverage the compatibility function to recognize novel signs at test time. 

The performance of the resulting zero-shot sign recognition model directly depends on three factors: (i) video
representation, (ii) class representation, and, (iii) the model used as the compatibility function $F$. The following 
three sections provide the corresponding details.

\subsection{Spatio-temporal video embedding}
We aim to obtain an effective video representation by extracting short-term spatio-temporal features using ConvNet
features of the video snippets, and then capturing longer-term dynamics through recurrent models.
We additionally improve our representation by extracting features in two separate streams: the full frames and
hand regions only. The details are given in the following paragraphs.

\myparagraphwospace{Short-term spatio-temporal representation.}
We obtain our basic spatio-temporal representation by first splitting each video into 8 frames long snippets
and then extracting their features using a pre-trained I3D model~\cite{carreira2017quo}, a state-of-the-art 
3D-ConvNet architecture. The I3D model is obtained by adapting a pre-trained Inception model \cite{szegedy2015going} to the 
video domain and then fine-tuning on the Kinetics dataset. We obtain our most basic video representation by
average pooling the resulting snippet features.

\myparagraphwospace{Modeling longer-term dependencies.}
Average pooling the 3D-CNN features is a well-performing technique for the recognition of 
non-complex (singleton) actions. Signs, on the contrary, portray more complicated 
patterns that are composed of the sequences of multiple basic gestures. In order to capture the transition dynamics
and longer-term dependencies across the snippets of a video, we use recurrent network models that take the
I3D representation sequence as input, and, provide an output embedding. For this purpose, we propose to use
the bidirection LSTM (bi-LSTM)~\cite{graves2005framewise} model, and, compare it against the average pooling,
LSTM~\cite{hochreiter1997long} and GRU~\cite{cho2014learning} models.

\myparagraphwospace{Two-stream video representation.}
Hands play a central role in expressing signs. In order to encode details of the hand-area information in a manner
isolated from the the overall body movements,
we detect and crop the hand regions using OpenPose \cite{cao2018openpose} and form a hand-only sequence
corresponding to each video snippet. We define two separate streams, including I3D and bi-LSTM networks, over these video inputs and then concatenate the resulting features to obtain the final video representation (Figure~\ref{fig:main_figure}).
When using recurrent networks, we train both streams together with the compatibility function in an end-to-end fashion.

    \begin{figure}
      \centering
          \includegraphics[width=\textwidth]{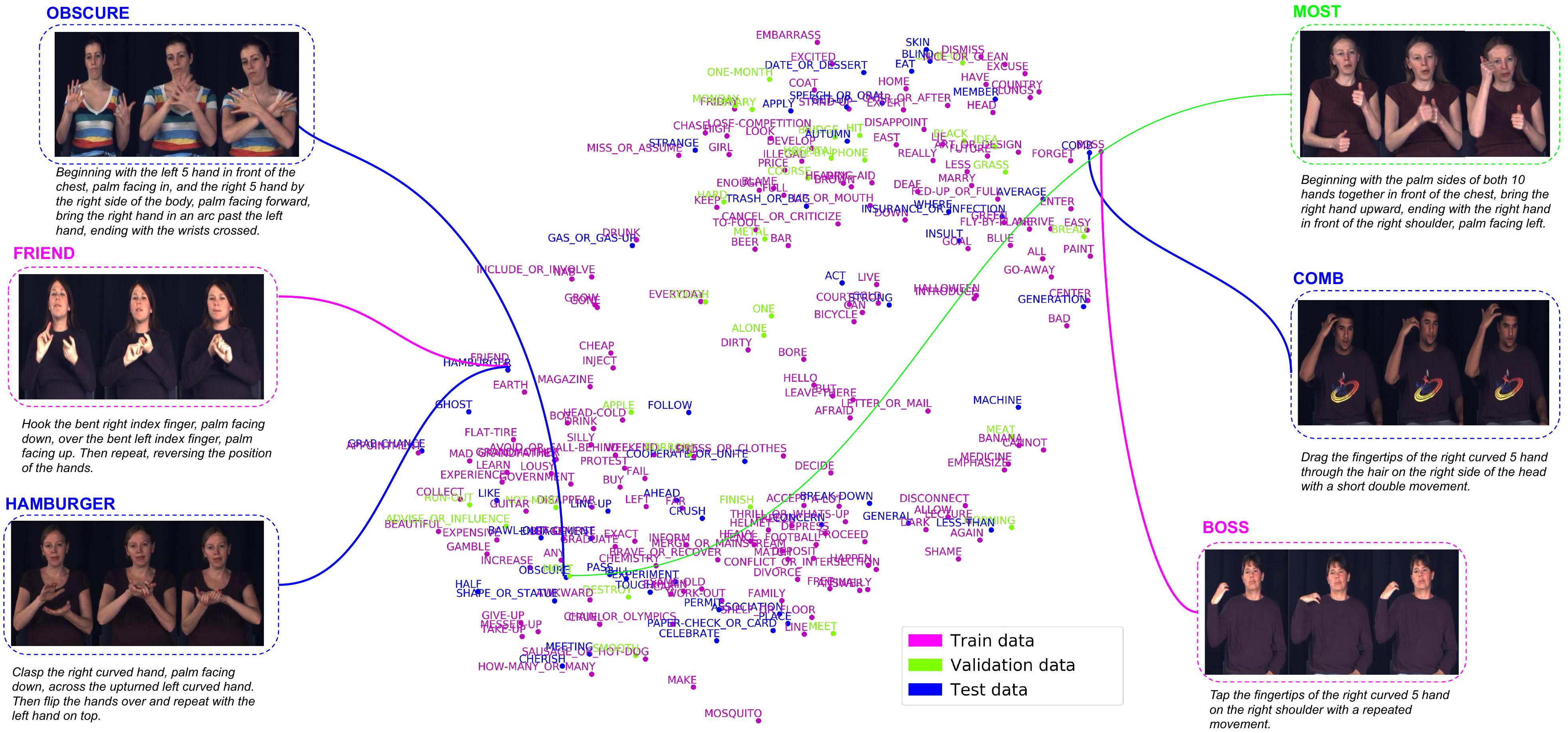}
          \vspace{-.2cm}
            \caption{t-SNE visualization of sign descriptions using BERT-\cite{devlin2018bert} embeddings. Nearby descriptions typically correspond to visually similar signs. Best viewed in color, with zoom.}
                \label{fig:BERT}
          \vspace{-.5cm}
    \end{figure}

\subsection{Text-based class embeddings}
We extract contextualized language embeddings from textual sign descriptions using the state-of-the-art language representation model BERT~\cite{devlin2018bert}. BERT architecture basically consists of a stack of encoders; specifically, multi-layer bidirectional transformers\cite{vaswani2017attention}. The model's main advantage over word2vec \cite{mikolov2013distributed} and glove \cite{pennington2014glove} representations is that BERT model is contextual and the extracted representations of the words change with respect to other words in a sentence. 

Figure \ref{fig:BERT} shows the t-SNE visualization of all sign class BERT embeddings. A close inspection to this feature space reveals that classes that appear closer in t-SNE embeddings
have indeed similar descriptions.
For instance, \textit{friend} and \textit{hamburger} signs are composed of similar motions with different hand-shapes,
\textit{obscure} and \textit{most} signs have similar hand movements but different hand shapes and directions,
and,
\textit{comb} and \textit{boss} signs include the same repeated movement with different hand-shape and locations with respect to the body.

\vspace{-3mm}

\subsection{Zero-shot learning model}
In our work, we adapt a label embedding~\cite{akata2013label,weston2010large} based formulation to tackle the ZSSLR
problem. More specifically, we use bi-linear compatibility function that associates the video and class representations:
\begin{equation}
    F(v,c) = \theta(v)^T W \phi(\tau(c))
\end{equation}
where $\theta(v)$ is the $d$-dimensional embedding of the video $v$,
$\phi(\tau(c))$ is the $m$-dimensional BERT embeddings of the textual description $\tau(c)$ for the class $c$, and, $W$ is the $d{\times}m$ compatibility matrix. In order to learn this matrix, we use the cross entropy loss with $\ell_2$-regularization:
\begin{equation}
    \min_{W} - \dfrac{1}{N} \sum_{i=1}^{N} \log \frac{\exp \{ F(v_i,c_i) \} }{ \sum_{c_j \in \clss} \exp \{ F(v_i,c_j) \} } + \lambda \| W \|^2
\end{equation}
where $\lambda$ is the regularization weight. This core formulation is also used in \cite{sumbul2018fine}, in a completely different ZSL problem. Since 
the objective function is analogous to the logistic regression classifier, we refer to this approach as {\em logistic label embedding} (LLE).

In addition to LLE, we also adapt the {\em embarrassingly simple zero-shot learning} (ESZSL)~\cite{romera2015embarrassingly} and {\em semantic auto-encoder} (SAE)~\cite{kodirov2017semantic} formulations as baselines. We, however, skip their formulational details here for brevity.

\vspace{-3mm}

\section{Experiments}

\subsection{Implementation Details}
\label{sec:implementationdetails}
 We fix the number of video frames of each sign video to 32 by either down-scaling or up-scaling. For every consecutive 8 frames, we extract 1024-d features from the last average pooling layer of the I3D model  using a stride of 4. When modeling the longer temporal context, we set LSTM's or bi-LSTM's initial hidden and cell state to average pool of each sequence during training. Hence, hidden size equals to the size of average pooled feature vector, which is 1024. For representing text, we use the $\mathrm{BERT_{BASE}}$ model \cite{devlin2018bert} and extract 768-dimensional sentence-based features. Following the description in \cite{devlin2018bert}, we concatenate the features from the last four layers of the pretrained Transformer of $\mathrm{BERT_{BASE}}$ and $l_2$-normalize them. 
 
 We measure normalized accuracy, \ie the mean accuracy per class, in all experiments. We run each experiment 5 times and report the average.  Top-1, top-2, and top-5 accuracies for the random baseline are calculated by averaging over 10000 runs.

\begin{table}[t]
\centering
\caption{Comparison of different ZSL formulations. Here, I3D~\cite{carreira2017quo} features are extracted over the whole frames (\ie body stream) only.}
\label{fig:res1}
\small{
\begin{tabular}{lllll}
\toprule
Method & Val (30 Classes) & \multicolumn{3}{c}{Test (50 Classes)} \\
       & top- 1                 & top-1 & top-2 & top-5 \\
\midrule
Random  & 3.3                    &  2.0 & 4.0 & 10.0  \\ \hline
SAE  & 10.6 &  8.0 & 12.0 & 16.0  \\ 
ESZSL  & 12.0  & \textbf{16.9} & \textbf{26.0} & \textbf{44.4}  \\ 
LLE      &  \textbf{14.1} & 11.4 & 21.2 & 41.1 \\ 

\bottomrule
\end{tabular}
}
\vspace{-1mm}
\end{table}

\vspace{-3mm}
\subsection{Experimental Results}

We first evaluate the ZSL component of our framework. In this context, we explore three different ZSL approaches, namely 
SAE \cite{kodirov2017semantic}, ESZSL \cite{romera2015embarrassingly}, and LLE. In these set of experiments, we have pooled the extracted 3D-CNN features over the whole frame. Table \ref{fig:res1} shows the corresponding results, where top-1 validation accuracy, and top-1, top-2, and top-5 test accuracies are reported. 

We observe that SAE \cite{kodirov2017semantic} performs poorly with respect to other approaches. We think that this is due to auto-encoder structure of SAE method. The model learns linear embedding from video to semantic space with the purpose of 
reconstruction back from semantic space to video. This idea might not work well since we do not have many in-class
samples for reconstruction. In addition, as stated earlier, intra-class variance is very high among signers.

Consequently, we evaluate the performance of the two-stream  spatio-temporal representation of the framework. Specifically, we carry out an ablation study, where \textit{body} denotes the full frame input stream, \textit{hand} denotes the hand videos and \textit{body+hand} is the case when these two streams are used in conjunction. 
The corresponding results are given in Table \ref{fig:res2}. Hand stream  provides additional cues and increases the performance for validation classes using both methods. In test classes, ESZSL \cite{romera2015embarrassingly} does not perform well on the hand-stream; on the contrary, its performance increases when both streams are used in conjunction. Similarly, LLE benefits from the introduction of hand-stream, and outperforms ESZSL method when two streams are utilized together. Overall, we observe that proposed framework based on LLE formulation works better, especially regarding top-1 and top-2 accuracies.

\begin{table}[t]
\centering
\caption{Evaluation of two-stream spatio-temporal representation. Here, \textit{body} denotes the full frame input stream, whereas \textit{hand} denotes the videos of hand regions and \textit{body+hand} is the case when these two streams are used in conjunction. Here, average pooling is used in aggregating the short-term video representations. }
\label{fig:res2}
\small{
\begin{tabular}{llllll}
\toprule
Method & visual rep. & Val (30 Classes) & \multicolumn{3}{c}{Test (50 Classes)} \\
       &       & top - 1                 & top-1 & top-2 & top-5 \\
\midrule
Random                     & \hspace{0.2cm} -    &  3.3  & 2.0 & 4.0 & 10.0  \\ \hline
\multirow{3}{*}{ESZSL} &  body &  12.0  & 16.9 & 26.0 & \textbf{44.4}\\
                            & hand &  13.3  & 11.6 & 19.6 & 33.7 \\
                           & body + hand &  14.6  &  17.1 & 25.7 & 43.0 \\ \hline

\multirow{3}{*}{LLE} & body & 14.1   & 11.4 & 21.2 & 41.1 \\
                            & hand &  15.0  & 12.6 & 19.8 & 37.8 \\
                           & body + hand &  \textbf{16.2}   &  \textbf{18.0} & \textbf{27.4} & 43.8 \\ 
\bottomrule
\end{tabular}
}
\vspace{-2mm}
\end{table}

\begin{table}[]
\centering
\caption{Comparison of different RNN units with LLE method.}
\label{fig:res3}
\small{
\begin{tabular}{llll}
\toprule
Temporal Representation & top-1 & top-2 & top-5 \\
\midrule
AvePool    & 18.0  & 27.4  & 43.8     \\
                           LSTM  \cite{hochreiter1997long}    &  18.2 & 28  & 47.2  \\
                           GRU \cite{cho2014learning}       & 19.7  & 31.8  & 50.0  \\
                           bi-LSTM \cite{graves2005framewise}  &  \textbf{20.9} & \textbf{32.5}  & \textbf{51.4}  \\

\bottomrule
\end{tabular}
}
\end{table}

We further evaluate the effect of longer temporal modeling with different RNN architectures. We experiment with three different RNN  models, namely LSTM\cite{hochreiter1997long}, GRU\cite{cho2014learning} and bi-LSTM\cite{graves2005framewise} units using LLE over both hand and body streams. Table \ref{fig:res3} presents these results. We observe that, compared to average pooling of streams, the framework benefits from the introduction of longer temporal modeling over all architectures, and performs the best with bi-LSTMs. This illustrates the importance of visual representations for ZSSLR. Our overall proposed framework reaches a top-1 normalized accuracy of 20.9\%  and top-5 normalized accuracy of 51.4\%, which is quite impressive compared to top-1 and top-5 accuracies of random baseline (2.0\% and 10.0\% respectively).

\begin{figure*}[!t]
\centering
\begin{tabular}{>{\raggedright\arraybackslash}m{0.35\linewidth}>{\raggedright\arraybackslash}m{0.6\linewidth}}
\toprule
\includegraphics[width=0.3\linewidth, height=1.8cm]{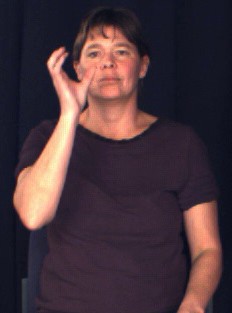}
\includegraphics[width=0.3\linewidth, height=1.8cm]{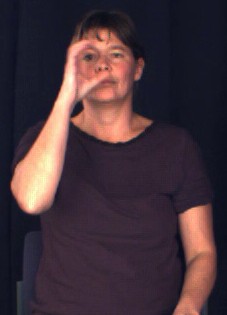}
\includegraphics[width=0.3\linewidth, height=1.8cm]{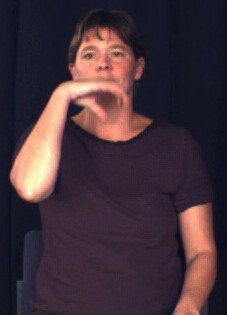} 
&
\vspace{-5mm}

\begin{minipage}{0.6\textwidth}
\scriptsize{\textbf{\\ \\Correctly Predicted Label: STRANGE}}\\
\scriptsize{Move the right \textcolor{red}{C} hand from near the right side of the face, palm facing left, downward in an arc in front of the face, ending near the left side of the chin, palm facing down.}
\end{minipage}
\\
\includegraphics[width=0.3\linewidth, height=1.8cm]{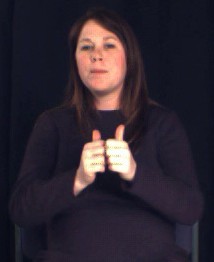}
\includegraphics[width=0.3\linewidth, height=1.8cm]{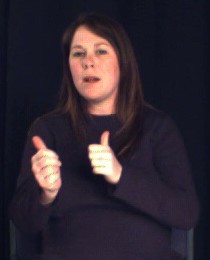}
\includegraphics[width=0.3\linewidth, height=1.8cm]{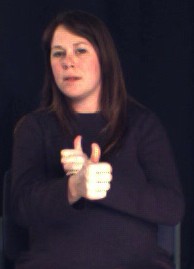} 
 & 
 \vspace{-5mm}
\begin{minipage}{0.6\textwidth}
\scriptsize{
\textbf{\\ \\  \\ Correctly Predicted Label: AHEAD}\\
Beginning with the palm sides of both \textcolor{red}{A} hands together, move the right hand forward in a small arc.}
\end{minipage} 

\\
\includegraphics[width=0.3\linewidth, height=1.8cm]{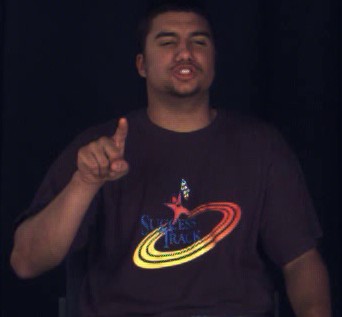}
\includegraphics[width=0.3\linewidth, height=1.8cm]{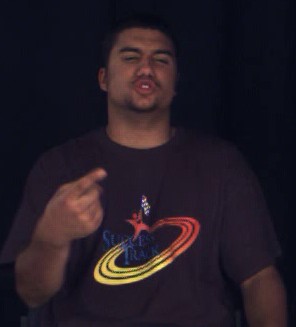}
\includegraphics[width=0.3\linewidth, height=1.8cm]{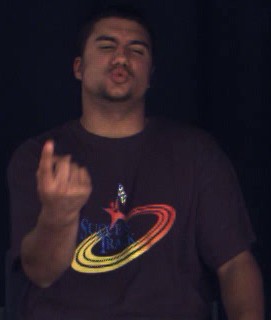} 
 & 
 \vspace{-5mm}
\begin{minipage}{0.6\textwidth}
\scriptsize{ \textbf{\\ \\ Correctly Predicted Label: INSULT}\\
Move the extended right index finger from in front of the right side of the body, palm facing left and finger pointing forward, forward and upward sharply in an arc.}
\end{minipage} 
\\
\includegraphics[width=0.3\linewidth, height=1.8cm]{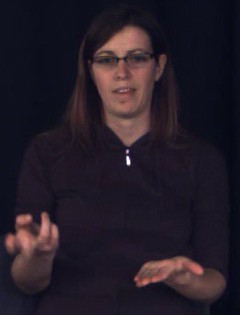}
\includegraphics[width=0.3\linewidth, height=1.8cm]{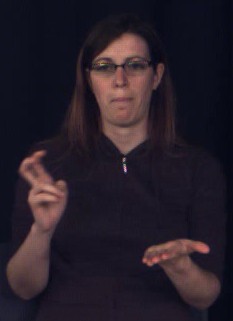}
\includegraphics[width=0.3\linewidth, height=1.8cm]{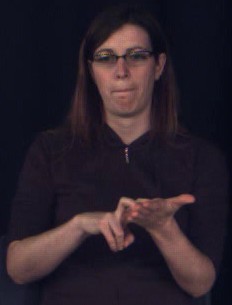} 
 & 
 \vspace{-5mm}
\begin{minipage}{0.6\textwidth}
\scriptsize{ \textbf{\\ \\ \\ \\Correctly Predicted Label: GET-UP}\\
Place the fingertips of the right bent \textcolor{red}{V} hand, palm facing in and fingers pointing down, on the upturned palm of the left open hand held in front of the body.}
\end{minipage} 

\\
\midrule
\includegraphics[width=0.3\linewidth, height=1.8cm]{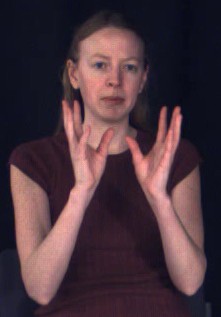}
\includegraphics[width=0.3\linewidth, height=1.8cm]{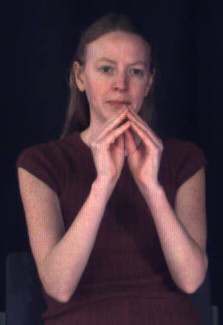}
\includegraphics[width=0.3\linewidth, height=1.8cm]{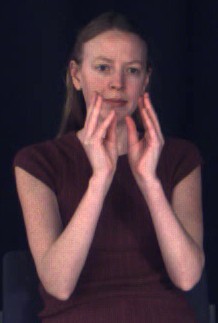}
 & 
\begin{minipage}{0.6\textwidth}
\scriptsize{\textbf{Predicted Label: BREAK-DOWN} \\
Beginning with the fingertips of both curved \textcolor{red}{5} hands touching in front of the chest, palms facing each other, allow the fingers to loosely drop, ending with the palms facing down.\\
\textbf{Correct Label: MEETING} \\
Beginning with both open hands in front of the chest, palms facing each other and fingers pointing up, close the fingers with a double movement into flattened \textcolor{red}{O} hands while moving the hands together.
}
\end{minipage}
\\
\\
\includegraphics[width=0.3\linewidth, height=1.8cm]{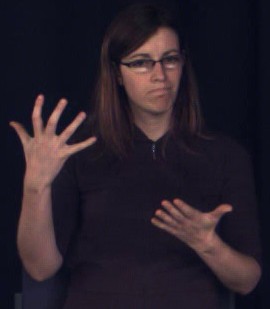}
\includegraphics[width=0.3\linewidth, height=1.8cm]{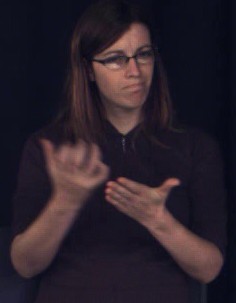}
\includegraphics[width=0.3\linewidth, height=1.8cm]{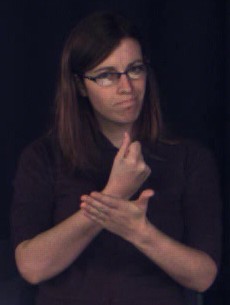}
& 
\begin{minipage}{0.6\textwidth}
\scriptsize{\textbf{Predicted Label: AVERAGE} \\
Brush the little-finger side of the right open hand, palm facing left, back and forth with a short repeated movement on the index-finger side of the left open hand, palm angled right.\\
\textbf{Correct Label: GRAB-CHANCE} \\
Bring the right curved \textcolor{red}{5} hand from in front of the right side of the body, palm facing left and fingers pointing forward, in toward the body in a downward arc while changing into an S hand, brushing the little-finger side of the right \textcolor{red}{S} hand across the palm of the left open hand, palm facing up in front of the chest.
}
\end{minipage} 
\\
\bottomrule

\end{tabular}
\caption{Example predictions of our proposed model. The first four rows show examples that are correctly predicted and the last two rows show incorrect predictions, together with the textual descriptions of the predicted and ground-truth classes.}
\label{fig:prediction}
\vspace{-5mm}
\end{figure*}

Figure~\ref{fig:prediction} shows examples from correctly and incorrectly classified test sequences. We observe that, either the textual descriptions or the visual aspects of the classes confused with each other are very similar. This indicates that the problem domain can benefit from more detailed analyses and representations that focus on nuances, both in visual and in textual domain, which can be explored as a future direction.

\vspace{-3mm}

\section{Conclusion}
This paper introduces and explores the problem of ZSSLR. We present a benchmark dataset for this novel problem by augmenting a large ASL dataset with sign language dictionary descriptions. Our proposed framework builds upon the idea of using these auxiliary texts as an additional source of information to recognize unseen signs. We propose an end-to-end trainable ZSSLR method that focuses hand and full body regions via 3D-CNNs+LSTMs and learns a compatibility function via label embedding. Overall, the experimental results indicate that, zero-shot recognition of signs based on textual descriptions can be possible. Nevertheless, the acquired accuracy levels are quite low compared to other ZSL domains, pinpointing a substantial need for further exploration in this direction.   
\section{Acknowledgements}
This work was supported in part by TUBITAK Career Grant 116E445.

{\small
\bibliography{egbib}}

\begin{thebibliography}{67}
\providecommand{\natexlab}[1]{#1}
\providecommand{\url}[1]{\texttt{#1}}
\expandafter\ifx\csname urlstyle\endcsname\relax
  \providecommand{\doi}[1]{doi: #1}\else
  \providecommand{\doi}{doi: \begingroup \urlstyle{rm}\Url}\fi

\bibitem[Akata et~al.(2013)Akata, Perronnin, Harchaoui, and
  Schmid]{akata2013label}
Zeynep Akata, Florent Perronnin, Zaid Harchaoui, and Cordelia Schmid.
\newblock Label-embedding for attribute-based classification.
\newblock In \emph{Proc. IEEE Conf. Comput. Vis. Pattern Recog.}, pages
  819--826, 2013.

\bibitem[Buehler et~al.(2009)Buehler, Zisserman, and
  Everingham]{buehler2009learning}
Patrick Buehler, Andrew Zisserman, and Mark Everingham.
\newblock Learning sign language by watching tv (using weakly aligned
  subtitles).
\newblock In \emph{Proc. IEEE Conf. Comput. Vis. Pattern Recog.}, pages
  2961--2968. IEEE, 2009.

\bibitem[Camgoz et~al.(2017)Camgoz, Hadfield, Koller, and
  Bowden]{camgoz2017subunets}
Necati~Cihan Camgoz, Simon Hadfield, Oscar Koller, and Richard Bowden.
\newblock Subunets: End-to-end hand shape and continuous sign language
  recognition.
\newblock In \emph{Proc. IEEE Int. Conf. on Computer Vision}, pages 3075--3084.
  IEEE, 2017.

\bibitem[Cao et~al.(2018)Cao, Hidalgo, Simon, Wei, and Sheikh]{cao2018openpose}
Zhe Cao, Gines Hidalgo, Tomas Simon, Shih-En Wei, and Yaser Sheikh.
\newblock Open{P}ose: realtime multi-person 2{D} pose estimation using {P}art
  {A}ffinity {F}ields.
\newblock In \emph{arXiv preprint arXiv:1812.08008}, 2018.

\bibitem[Carreira and Zisserman(2017)]{carreira2017quo}
Joao Carreira and Andrew Zisserman.
\newblock Quo vadis, action recognition? a new model and the kinetics dataset.
\newblock In \emph{Proc. IEEE Conf. Comput. Vis. Pattern Recog.}, pages
  6299--6308, 2017.

\bibitem[Cho et~al.(2014)Cho, Van~Merri{\"e}nboer, Gulcehre, Bahdanau,
  Bougares, Schwenk, and Bengio]{cho2014learning}
Kyunghyun Cho, Bart Van~Merri{\"e}nboer, Caglar Gulcehre, Dzmitry Bahdanau,
  Fethi Bougares, Holger Schwenk, and Yoshua Bengio.
\newblock Learning phrase representations using rnn encoder-decoder for
  statistical machine translation.
\newblock \emph{EMNLP}, 2014.

\bibitem[Cihan~Camgoz et~al.(2018)Cihan~Camgoz, Hadfield, Koller, Ney, and
  Bowden]{cihan2018neural}
Necati Cihan~Camgoz, Simon Hadfield, Oscar Koller, Hermann Ney, and Richard
  Bowden.
\newblock Neural sign language translation.
\newblock In \emph{Proc. IEEE Conf. Comput. Vis. Pattern Recog.}, pages
  7784--7793, 2018.

\bibitem[Costello(1999)]{costello1999random}
Elaine Costello.
\newblock \emph{Random House Webster's Concise American Sign Language
  Dictionary}.
\newblock Random House, 1999.

\bibitem[Cui et~al.(2017)Cui, Liu, and Zhang]{cui2017recurrent}
Runpeng Cui, Hu~Liu, and Changshui Zhang.
\newblock Recurrent convolutional neural networks for continuous sign language
  recognition by staged optimization.
\newblock In \emph{Proc. IEEE Conf. Comput. Vis. Pattern Recog.}, pages
  7361--7369, 2017.

\bibitem[Devlin et~al.(2018)Devlin, Chang, Lee, and Toutanova]{devlin2018bert}
Jacob Devlin, Ming-Wei Chang, Kenton Lee, and Kristina Toutanova.
\newblock Bert: Pre-training of deep bidirectional transformers for language
  understanding.
\newblock \emph{arXiv preprint arXiv:1810.04805}, 2018.

\bibitem[Elhoseiny et~al.(2013)Elhoseiny, Saleh, and
  Elgammal]{elhoseiny2013write}
Mohamed Elhoseiny, Babak Saleh, and Ahmed Elgammal.
\newblock Write a classifier: Zero-shot learning using purely textual
  descriptions.
\newblock In \emph{Proc. IEEE Int. Conf. on Computer Vision}, pages 2584--2591,
  2013.

\bibitem[Farhadi and Forsyth(2006)]{farhadi2006aligning}
Ali Farhadi and David Forsyth.
\newblock Aligning asl for statistical translation using a discriminative word
  model.
\newblock In \emph{Proc. IEEE Conf. Comput. Vis. Pattern Recog.}, volume~2,
  pages 1471--1476. IEEE, 2006.

\bibitem[Farhadi et~al.(2007)Farhadi, Forsyth, and White]{farhadi2007transfer}
Ali Farhadi, David Forsyth, and Ryan White.
\newblock Transfer learning in sign language.
\newblock In \emph{Proc. IEEE Conf. Comput. Vis. Pattern Recog.}, pages 1--8.
  IEEE, 2007.

\bibitem[Farhadi et~al.(2009)Farhadi, Endres, Hoiem, and
  Forsyth]{farhadi2009describing}
Ali Farhadi, Ian Endres, Derek Hoiem, and David Forsyth.
\newblock Describing objects by their attributes.
\newblock In \emph{Proc. IEEE Conf. Comput. Vis. Pattern Recog.}, pages
  1778--1785. IEEE, 2009.

\bibitem[Ferrari and Zisserman(2008)]{ferrari2008learning}
Vittorio Ferrari and Andrew Zisserman.
\newblock Learning visual attributes.
\newblock In \emph{Proc. Adv. Neural Inf. Process. Syst.}, pages 433--440,
  2008.

\bibitem[Frome et~al.(2013)Frome, Corrado, Shlens, Bengio, Dean, Mikolov,
  et~al.]{frome2013devise}
Andrea Frome, Greg~S Corrado, Jon Shlens, Samy Bengio, Jeff Dean, Tomas
  Mikolov, et~al.
\newblock Devise: A deep visual-semantic embedding model.
\newblock In \emph{Proc. Adv. Neural Inf. Process. Syst.}, pages 2121--2129,
  2013.

\bibitem[Fu et~al.(2014{\natexlab{a}})Fu, Hospedales, Xiang, Fu, and
  Gong]{fu2014transductive}
Yanwei Fu, Timothy~M Hospedales, Tao Xiang, Zhenyong Fu, and Shaogang Gong.
\newblock Transductive multi-view embedding for zero-shot recognition and
  annotation.
\newblock In \emph{Proc. European Conf. on Computer Vision}, pages 584--599.
  Springer, 2014{\natexlab{a}}.

\bibitem[Fu et~al.(2014{\natexlab{b}})Fu, Hospedales, Xiang, and
  Gong]{fu2014learning}
Yanwei Fu, Timothy~M Hospedales, Tao Xiang, and Shaogang Gong.
\newblock Learning multimodal latent attributes.
\newblock \emph{IEEE transactions on pattern analysis and machine
  intelligence}, 36\penalty0 (2):\penalty0 303--316, 2014{\natexlab{b}}.

\bibitem[Graves and Schmidhuber(2005)]{graves2005framewise}
Alex Graves and J{\"u}rgen Schmidhuber.
\newblock Framewise phoneme classification with bidirectional lstm and other
  neural network architectures.
\newblock \emph{Neural Networks}, 18\penalty0 (5-6):\penalty0 602--610, 2005.

\bibitem[Grobel and Assan(1997)]{grobel1997isolated}
Kirsti Grobel and Marcell Assan.
\newblock Isolated sign language recognition using hidden markov models.
\newblock In \emph{IEEE International Conference on Systems, Man, and
  Cybernetics. Computational Cybernetics and Simulation}, volume~1, pages
  162--167. IEEE, 1997.

\bibitem[Habibian et~al.(2017)Habibian, Mensink, and
  Snoek]{habibian2017video2vec}
Amirhossein Habibian, Thomas Mensink, and Cees~GM Snoek.
\newblock Video2vec embeddings recognize events when examples are scarce.
\newblock \emph{IEEE Trans. Pattern Anal. Mach. Intell.}, 39\penalty0
  (10):\penalty0 2089--2103, 2017.

\bibitem[Hochreiter and Schmidhuber(1997)]{hochreiter1997long}
Sepp Hochreiter and J{\"u}rgen Schmidhuber.
\newblock Long short-term memory.
\newblock \emph{Neural computation}, 9\penalty0 (8):\penalty0 1735--1780, 1997.

\bibitem[Huang and Huang(1998)]{huang1998sign}
Chung-Lin Huang and Wen-Yi Huang.
\newblock Sign language recognition using model-based tracking and a 3d
  hopfield neural network.
\newblock \emph{Machine vision and applications}, 10\penalty0 (5-6):\penalty0
  292--307, 1998.

\bibitem[Huang et~al.(2015)Huang, Zhou, Li, and Li]{huang2015sign}
Jie Huang, Wengang Zhou, Houqiang Li, and Weiping Li.
\newblock Sign language recognition using 3d convolutional neural networks.
\newblock In \emph{IEEE {I}nt. {C}onf. on {M}ultimedia and {E}xpo (ICME)},
  pages 1--6. IEEE, 2015.

\bibitem[Jain et~al.(2015)Jain, van Gemert, Mensink, and
  Snoek]{jain2015objects2action}
Mihir Jain, Jan~C van Gemert, Thomas Mensink, and Cees~GM Snoek.
\newblock Objects2action: Classifying and localizing actions without any video
  example.
\newblock In \emph{Proc. IEEE Int. Conf. on Computer Vision}, pages 4588--4596,
  2015.

\bibitem[Kelly et~al.(2011)Kelly, Mc~Donald, and Markham]{kelly2011weakly}
Daniel Kelly, John Mc~Donald, and Charles Markham.
\newblock Weakly supervised training of a sign language recognition system
  using multiple instance learning density matrices.
\newblock \emph{IEEE Transactions on Systems, Man, and Cybernetics, Part B
  (Cybernetics)}, 41\penalty0 (2):\penalty0 526--541, 2011.

\bibitem[Kodirov et~al.(2017)Kodirov, Xiang, and Gong]{kodirov2017semantic}
Elyor Kodirov, Tao Xiang, and Shaogang Gong.
\newblock Semantic autoencoder for zero-shot learning.
\newblock In \emph{Proc. IEEE Conf. Comput. Vis. Pattern Recog.}, pages
  3174--3183, 2017.

\bibitem[Koller et~al.(2015)Koller, Forster, and Ney]{koller2015continuous}
Oscar Koller, Jens Forster, and Hermann Ney.
\newblock Continuous sign language recognition: Towards large vocabulary
  statistical recognition systems handling multiple signers.
\newblock \emph{Comput. Vis. Image Understand.}, 141:\penalty0 108--125, 2015.

\bibitem[Koller et~al.(2016)Koller, Zargaran, Ney, and Bowden]{koller2016deep}
Oscar Koller, O~Zargaran, Hermann Ney, and Richard Bowden.
\newblock Deep sign: hybrid cnn-hmm for continuous sign language recognition.
\newblock In \emph{British Machine Vision Conference}, 2016.

\bibitem[Lampert et~al.(2009)Lampert, Nickisch, and
  Harmeling]{lampert2009learning}
Christoph~H Lampert, Hannes Nickisch, and Stefan Harmeling.
\newblock Learning to detect unseen object classes by between-class attribute
  transfer.
\newblock In \emph{Proc. IEEE Conf. Comput. Vis. Pattern Recog.}, pages
  951--958. IEEE, 2009.

\bibitem[Lampert et~al.(2014)Lampert, Nickisch, and
  Harmeling]{lampert2014attribute}
Christoph~H Lampert, Hannes Nickisch, and Stefan Harmeling.
\newblock Attribute-based classification for zero-shot visual object
  categorization.
\newblock \emph{IEEE Transactions on Pattern Analysis and Machine
  Intelligence}, 36\penalty0 (3):\penalty0 453--465, 2014.

\bibitem[Lei~Ba et~al.(2015)Lei~Ba, Swersky, Fidler, et~al.]{lei2015predicting}
Jimmy Lei~Ba, Kevin Swersky, Sanja Fidler, et~al.
\newblock Predicting deep zero-shot convolutional neural networks using textual
  descriptions.
\newblock In \emph{Proc. IEEE Int. Conf. on Computer Vision}, pages 4247--4255,
  2015.

\bibitem[Liu et~al.(2011)Liu, Kuipers, and Savarese]{liu2011recognizing}
Jingen Liu, Benjamin Kuipers, and Silvio Savarese.
\newblock Recognizing human actions by attributes.
\newblock In \emph{Proc. IEEE Conf. Comput. Vis. Pattern Recog.}, pages
  3337--3344. IEEE, 2011.

\bibitem[Madapana and Wachs(2018)]{madapana2018hard}
Naveen Madapana and Juan~P Wachs.
\newblock Hard zero shot learning for gesture recognition.
\newblock In \emph{IAPR International Conference on Pattern Recognition}, pages
  3574--3579. IEEE, 2018.

\bibitem[Mensink et~al.(2014)Mensink, Gavves, and Snoek]{mensink2014costa}
Thomas Mensink, Efstratios Gavves, and Cees~GM Snoek.
\newblock Costa: Co-occurrence statistics for zero-shot classification.
\newblock In \emph{Proc. IEEE Conf. Comput. Vis. Pattern Recog.}, pages
  2441--2448, 2014.

\bibitem[Mikolov et~al.(2013)Mikolov, Sutskever, Chen, Corrado, and
  Dean]{mikolov2013distributed}
Tomas Mikolov, Ilya Sutskever, Kai Chen, Greg~S Corrado, and Jeff Dean.
\newblock Distributed representations of words and phrases and their
  compositionality.
\newblock In \emph{Proc. Adv. Neural Inf. Process. Syst.}, pages 3111--3119,
  2013.

\bibitem[Molchanov et~al.(2016)Molchanov, Yang, Gupta, Kim, Tyree, and
  Kautz]{molchanov2016online}
Pavlo Molchanov, Xiaodong Yang, Shalini Gupta, Kihwan Kim, Stephen Tyree, and
  Jan Kautz.
\newblock Online detection and classification of dynamic hand gestures with
  recurrent 3d convolutional neural network.
\newblock In \emph{Proc. IEEE Conf. Comput. Vis. Pattern Recog.}, pages
  4207--4215, 2016.

\bibitem[Narayana et~al.(2018)Narayana, Beveridge, and
  Draper]{narayana2018gesture}
Pradyumna Narayana, Ross Beveridge, and Bruce~A Draper.
\newblock Gesture recognition: Focus on the hands.
\newblock In \emph{Proc. IEEE Conf. Comput. Vis. Pattern Recog.}, pages
  5235--5244, 2018.

\bibitem[Nayak et~al.(2009)Nayak, Sarkar, and Loeding]{nayak2009automated}
Sunita Nayak, Sudeep Sarkar, and Barbara Loeding.
\newblock Automated extraction of signs from continuous sign language sentences
  using iterated conditional modes.
\newblock In \emph{Proc. IEEE Conf. Comput. Vis. Pattern Recog.}, pages
  2583--2590. IEEE, 2009.

\bibitem[Neidle et~al.(2012)Neidle, Thangali, and
  Sclaroff]{neidle2012challenges}
Carol Neidle, Ashwin Thangali, and Stan Sclaroff.
\newblock Challenges in development of the american sign language lexicon video
  dataset (asllvd) corpus.
\newblock In \emph{Proc. 5th Workshop on the Representation and Processing of
  Sign Languages: Interactions between Corpus and Lexicon, Language Resources
  and Evaluation Conference (LREC) 2012}, 2012.

\bibitem[Norouzi et~al.(2014)Norouzi, Mikolov, Bengio, Singer, Shlens, Frome,
  Corrado, and Dean]{norouzi2013zero}
Mohammad Norouzi, Tomas Mikolov, Samy Bengio, Yoram Singer, Jonathon Shlens,
  Andrea Frome, Greg~S Corrado, and Jeffrey Dean.
\newblock Zero-shot learning by convex combination of semantic embeddings.
\newblock \emph{Proc. Int. Conf. Learn. Represent.}, 2014.

\bibitem[Parikh and Grauman(2011)]{parikh2011relative}
Devi Parikh and Kristen Grauman.
\newblock Relative attributes.
\newblock In \emph{Proc. IEEE Int. Conf. on Computer Vision}, pages 503--510.
  IEEE, 2011.

\bibitem[Patterson and Hays(2012)]{patterson2012sun}
Genevieve Patterson and James Hays.
\newblock Sun attribute database: Discovering, annotating, and recognizing
  scene attributes.
\newblock In \emph{Proc. IEEE Conf. Comput. Vis. Pattern Recog.}, pages
  2751--2758. IEEE, 2012.

\bibitem[Pennington et~al.(2014)Pennington, Socher, and
  Manning]{pennington2014glove}
Jeffrey Pennington, Richard Socher, and Christopher Manning.
\newblock Glove: Global vectors for word representation.
\newblock In \emph{Proc. of conference on empirical methods in natural language
  processing (EMNLP)}, pages 1532--1543, 2014.

\bibitem[Pfister et~al.(2013)Pfister, Charles, and Zisserman]{pfister2013large}
Tomas Pfister, James Charles, and Andrew Zisserman.
\newblock Large-scale learning of sign language by watching tv (using
  co-occurrences).
\newblock In \emph{British Machine Vision Conference}, 2013.

\bibitem[Pfister et~al.(2014)Pfister, Charles, and
  Zisserman]{pfister2014domain}
Tomas Pfister, James Charles, and Andrew Zisserman.
\newblock Domain-adaptive discriminative one-shot learning of gestures.
\newblock In \emph{Proc. European Conf. on Computer Vision}, pages 814--829.
  Springer, 2014.

\bibitem[Pigou et~al.(2016)Pigou, Van~Herreweghe, and Dambre]{pigou2016sign}
Lionel Pigou, Mieke Van~Herreweghe, and Joni Dambre.
\newblock Sign classification in sign language corpora with deep neural
  networks.
\newblock In \emph{International Conference on Language Resources and
  Evaluation (LREC) Workshop}, pages 175--178, 2016.

\bibitem[Qin et~al.(2017)Qin, Liu, Shao, Shen, Ni, Chen, and Wang]{qin2017zero}
Jie Qin, Li~Liu, Ling Shao, Fumin Shen, Bingbing Ni, Jiaxin Chen, and Yunhong
  Wang.
\newblock Zero-shot action recognition with error-correcting output codes.
\newblock In \emph{Proc. IEEE Conf. Comput. Vis. Pattern Recog.}, pages
  2833--2842, 2017.

\bibitem[Rohrbach et~al.(2011)Rohrbach, Stark, and
  Schiele]{rohrbach2011evaluating}
Marcus Rohrbach, Michael Stark, and Bernt Schiele.
\newblock Evaluating knowledge transfer and zero-shot learning in a large-scale
  setting.
\newblock In \emph{Proc. IEEE Conf. Comput. Vis. Pattern Recog.}, pages
  1641--1648. IEEE, 2011.

\bibitem[Romera-Paredes and Torr(2015)]{romera2015embarrassingly}
Bernardino Romera-Paredes and Philip Torr.
\newblock An embarrassingly simple approach to zero-shot learning.
\newblock In \emph{Proc. Int. Conf. Mach. Learn.}, pages 2152--2161, 2015.

\bibitem[Socher et~al.(2013)Socher, Ganjoo, Manning, and Ng]{socher2013zero}
Richard Socher, Milind Ganjoo, Christopher~D Manning, and Andrew Ng.
\newblock Zero-shot learning through cross-modal transfer.
\newblock In \emph{Proc. Adv. Neural Inf. Process. Syst.}, pages 935--943,
  2013.

\bibitem[Stokoe~Jr(2005)]{stokoe2005sign}
William~C Stokoe~Jr.
\newblock Sign language structure: An outline of the visual communication
  systems of the american deaf.
\newblock \emph{Journal of deaf studies and deaf education}, 10\penalty0
  (1):\penalty0 3--37, 2005.

\bibitem[Stoll et~al.(2018)Stoll, Camgoz, Hadfield, and Bowden]{stoll2018sign}
Stephanie Stoll, Necati~Cihan Camgoz, Simon Hadfield, and Richard Bowden.
\newblock Sign language production using neural machine translation and
  generative adversarial networks.
\newblock In \emph{British Machine Vision Conference}. British Machine Vision
  Association, 2018.

\bibitem[Sumbul et~al.(2018)Sumbul, Cinbis, and Aksoy]{sumbul2018fine}
Gencer Sumbul, Ramazan~Gokberk Cinbis, and Selim Aksoy.
\newblock Fine-grained object recognition and zero-shot learning in remote
  sensing imagery.
\newblock \emph{IEEE Transactions on Geoscience and Remote Sensing},
  56\penalty0 (2):\penalty0 770--779, 2018.

\bibitem[Szegedy et~al.(2015)Szegedy, Liu, Jia, Sermanet, Reed, Anguelov,
  Erhan, Vanhoucke, and Rabinovich]{szegedy2015going}
Christian Szegedy, Wei Liu, Yangqing Jia, Pierre Sermanet, Scott Reed, Dragomir
  Anguelov, Dumitru Erhan, Vincent Vanhoucke, and Andrew Rabinovich.
\newblock Going deeper with convolutions.
\newblock In \emph{Proc. IEEE Conf. Comput. Vis. Pattern Recog.}, pages 1--9,
  2015.

\bibitem[Tamura and Kawasaki(1988)]{tamura1988recognition}
Shinichi Tamura and Shingo Kawasaki.
\newblock Recognition of sign language motion images.
\newblock \emph{Pattern recognition}, 21\penalty0 (4):\penalty0 343--353, 1988.

\bibitem[Thomason and Knepper(2016)]{thomason2016recognizing}
Wil Thomason and Ross~A Knepper.
\newblock Recognizing unfamiliar gestures for human-robot interaction through
  zero-shot learning.
\newblock In \emph{International Symposium on Experimental Robotics}, pages
  841--852. Springer, 2016.

\bibitem[Vaswani et~al.(2017)Vaswani, Shazeer, Parmar, Uszkoreit, Jones, Gomez,
  Kaiser, and Polosukhin]{vaswani2017attention}
Ashish Vaswani, Noam Shazeer, Niki Parmar, Jakob Uszkoreit, Llion Jones,
  Aidan~N Gomez, {\L}ukasz Kaiser, and Illia Polosukhin.
\newblock Attention is all you need.
\newblock In \emph{Proc. Adv. Neural Inf. Process. Syst.}, pages 5998--6008,
  2017.

\bibitem[Wah et~al.(2011)Wah, Branson, Welinder, Perona, and
  Belongie]{wah2011caltech}
Catherine Wah, Steve Branson, Peter Welinder, Pietro Perona, and Serge
  Belongie.
\newblock The caltech-ucsd birds-200-2011 dataset.
\newblock 2011.

\bibitem[Wang et~al.(2016)Wang, Chai, Hong, Zhao, and Chen]{wang2016isolated}
Hanjie Wang, Xiujuan Chai, Xiaopeng Hong, Guoying Zhao, and Xilin Chen.
\newblock Isolated sign language recognition with grassmann covariance
  matrices.
\newblock \emph{ACM {T}ransactions on {A}ccessible {C}omputing (TACCESS)},
  8\penalty0 (4):\penalty0 14, 2016.

\bibitem[Wang and Chen(2017)]{wang2017alternative}
Qian Wang and Ke~Chen.
\newblock Alternative semantic representations for zero-shot human action
  recognition.
\newblock In \emph{Joint European Conference on Machine Learning and Knowledge
  Discovery in Databases}, pages 87--102. Springer, 2017.

\bibitem[Weston et~al.(2010)Weston, Bengio, and Usunier]{weston2010large}
Jason Weston, Samy Bengio, and Nicolas Usunier.
\newblock Large scale image annotation: learning to rank with joint word-image
  embeddings.
\newblock \emph{Machine learning}, 81\penalty0 (1):\penalty0 21--35, 2010.

\bibitem[Wu and Huang(1999)]{wu1999vision}
Ying Wu and Thomas~S Huang.
\newblock Vision-based gesture recognition: A review.
\newblock In \emph{International {G}esture {W}orkshop}, pages 103--115.
  Springer, 1999.

\bibitem[Xian et~al.(2017)Xian, Schiele, and Akata]{xian2017zero}
Yongqin Xian, Bernt Schiele, and Zeynep Akata.
\newblock Zero-shot learning-the good, the bad and the ugly.
\newblock In \emph{Proc. IEEE Conf. Comput. Vis. Pattern Recog.}, pages
  4582--4591, 2017.

\bibitem[Xu et~al.(2015)Xu, Hospedales, and Gong]{Xu2015SemanticES}
Xun Xu, Timothy~M. Hospedales, and Shaogang Gong.
\newblock Semantic embedding space for zero-shot action recognition.
\newblock \emph{2015 IEEE International Conference on Image Processing (ICIP)},
  pages 63--67, 2015.

\bibitem[Xu et~al.(2017)Xu, Hospedales, and Gong]{xu2017transductive}
Xun Xu, Timothy Hospedales, and Shaogang Gong.
\newblock Transductive zero-shot action recognition by word-vector embedding.
\newblock \emph{International Journal of Computer Vision}, 123\penalty0
  (3):\penalty0 309--333, 2017.

\bibitem[Zhu et~al.(2018)Zhu, Long, Guan, Newsam, and Shao]{zhu2018towards}
Yi~Zhu, Yang Long, Yu~Guan, Shawn Newsam, and Ling Shao.
\newblock Towards universal representation for unseen action recognition.
\newblock In \emph{Proc. IEEE Conf. Comput. Vis. Pattern Recog.}, pages
  9436--9445, 2018.

\end{thebibliography}
\end{document}